\documentclass[letterpaper]{article} % DO NOT CHANGE THIS
\usepackage{aaai24}  % DO NOT CHANGE THIS
\usepackage{times}  % DO NOT CHANGE THIS
\usepackage{helvet}  % DO NOT CHANGE THIS
\usepackage{courier}  % DO NOT CHANGE THIS
\usepackage[hyphens]{url}  % DO NOT CHANGE THIS
\usepackage{graphicx} % DO NOT CHANGE THIS
\usepackage{natbib}  % DO NOT CHANGE THIS AND DO NOT ADD ANY OPTIONS TO IT
\usepackage{caption} % DO NOT CHANGE THIS AND DO NOT ADD ANY OPTIONS TO IT
\usepackage{algorithm}
\usepackage{algorithmic}
\usepackage{subcaption}
\usepackage{color}
\usepackage{newfloat}
\usepackage{listings}
\DeclareCaptionStyle{ruled}{labelfont=normalfont,labelsep=colon,strut=off} % DO NOT CHANGE THIS
\floatstyle{ruled}
\newfloat{listing}{tb}{lst}{}
\floatname{listing}{Listing}
\usepackage{multirow}
\usepackage{amssymb}
\usepackage{amsmath}
\usepackage{xcolor}
\usepackage{xspace}
\usepackage{url}

\newcommand{\stitle}[1]{{\bf #1}}  % \noindent
\newcommand{\model}{{AMFormer}\xspace}
\newcommand{\modelauto}{{AMF-A}\xspace}
\newcommand{\modelftt}{{AMF-F}\xspace}
\newcommand{\ie}{\emph{i.e.,}\xspace}
\newcommand{\eg}{\emph{e.g.,}\xspace}
\newcommand{\eat}[1]{}

\usepackage{lipsum}

\title{Arithmetic Feature Interaction Is Necessary for Deep Tabular Learning}
\author {
    Yi Cheng\textsuperscript{\rm 123}\footnote{Equal contribution. Work done during Cheng's internship at Alibaba Group, under the guidance of Hu.},
    Renjun Hu\textsuperscript{\rm 4$\ast$},
    Haochao Ying\textsuperscript{\rm 135}\footnote{Corresponding author.},
    Xing Shi\textsuperscript{\rm 4},
    Jian Wu\textsuperscript{\rm 135},
    Wei Lin\textsuperscript{\rm 4}
}
\affiliations {
    \textsuperscript{\rm 1}State Key Laboratory of Transvascular Implantation Devices of the Second Affiliated Hospital, Zhejiang University School of Medicine, China\\
    \textsuperscript{\rm 2}School of Software Technology, Zhejiang University, China\\
    \textsuperscript{\rm 3}Institute of Wenzhou, Zhejiang University, China\\
    \textsuperscript{\rm 4}Alibaba Group\\
    \textsuperscript{\rm 5}School of Public Health, Zhejiang University, China\\
    
    \{chengy1, haochaoying, wujian2000\}@zju.edu.cn, \{renjun.hrj, shubao.sx, weilin.lw\}@alibaba-inc.com
}
\usepackage{bibentry}
\begin{document}

\maketitle

\begin{abstract}
Until recently, the question of the effective inductive bias of deep models on tabular data has remained unanswered. This paper investigates the hypothesis that arithmetic feature interaction is necessary for deep tabular learning. To test this point, we create a synthetic tabular dataset with a mild feature interaction assumption and examine a modified transformer architecture enabling arithmetical feature interactions, referred to as \model. Results show that \model outperforms strong counterparts in fine-grained tabular data modeling, data efficiency in training, and generalization. This is attributed to its parallel additive and multiplicative attention operators and prompt-based optimization, which facilitate the separation of tabular samples in an extended space with arithmetically-engineered features. Our extensive experiments on real-world data also validate the consistent effectiveness, efficiency, and rationale of \model, suggesting it has established a strong inductive bias for deep learning on tabular data. Code is available at \url{https://github.com/aigc-apps/AMFormer}.
\end{abstract}

\section{Introduction}
\label{sec:intro}

Tabular data is an extensively utilized and essential data format that finds its applications in diverse fields, including finance, marketing, medical science, and recommendation systems~\cite{Bank-Marketing, MIMIC-III, PhysioNet, MovieLens,Cheng2022EasyRecAE}. 
Such data often contains both categorical and numerical features, each of which holds its own specific meaning and relates to various modeling aspects. Due to the heterogeneity and potential sparsity of features, analyzing tabular data has remained a subject of research in the machine learning community. 
Among the many solutions proposed, tree ensemble models~\cite{xgboost,lightgbm,catboost} have emerged as the predominant choice, owing to their performance on various domains and robustness to data quality issues. 
Their success largely relies on the tree growth strategy, where each leaf exhaustively enumerates the splitting features and values, selecting the feature-value pair with the highest improvement on a certain criterion to divide the sample space. As a result, complex non-linear relationships between variables can be effectively captured. Meanwhile, since raw features directly involve, tree models often assume the features have been well-engineered~\cite{kddexpl2001}.

\eat{
Over the past few years, deep learning on tabular data has gained popularity as it eliminates the need for cumbersome and time-consuming feature engineering. Early attempts integrate deep neural networks (DNNs) to model high-order feature interactions~\cite{widendeep,deepfm,xdeepfm}, but this paradigm needs a good balance between model expressiveness and overfitting. 
To prevent overfitting and gain interpretability, a branch of studies has extended generalized additive models with DNNs to boost expressiveness in a more constrained manner~\cite{nam2021,nbm2022,sian2022}.
Additionally, tree-inspired architectures emulate the important elements of tree models using neural networks, allowing models to benefit from the strengths of both sides~\cite{node,Net-DNF}.
Finally, transformer architectures have also been explored for their effectiveness in capturing implicit feature interaction~\cite{autoint,fttrans}.
}
%, especially for sparse categorical features~\cite{kddexpl2001}.  ~\cite{GAM}
%, such as Wide \& Deep~\cite{widendeep} and deep factorization machines~\cite{deepfm,xdeepfm},
% NAM~\cite{nam2021}, NBM~\cite{nbm2022}, and SIAN~\cite{sian2022} 

In recent years, deep learning has become increasingly popular as a means to reduce the need for time-consuming and cumbersome feature engineering when dealing with tabular data. Early attempts at integrating deep neural networks (DNNs) aim to model high-order feature interactions~\cite{widendeep,deepfm,xdeepfm,Cheng2022EasyRecAE,DANet}; however, this paradigm requires a careful balance between model expressiveness and overfitting. To overcome this, researchers have turned to extending generalized additive models with DNNs to boost expressiveness in a more constrained manner~\cite{nam2021,nbm2022,sian2022,TabCaps}, thereby preventing overfitting and increasing interpretability. Others have explored tree-inspired architectures that emulate key elements of tree models using neural networks, taking advantage of the strengths of both techniques~\cite{node,Net-DNF}. Transformer has also been investigated for its success in the natural language and vision fields~\cite{autoint,fttrans,ExcelFormer,T2G-FORMER}.

%for their ability to effectively capture implicit feature interactions~\cite{autoint,fttrans}.

Despite multiple attempts, the effectiveness of deep learning on tabular data remains questionable~\cite{neuralranker} due to the unstable improvement over tree ensemble baselines, and tabular datasets have been considered as the last ``unconquered castle'' for deep learning~\cite{welltunednet}.
The central question is whether deep models have an effective inductive bias on tabular data. 
In this paper, we argue that arithmetic feature interaction is necessary for deep tabular learning.
More specifically, the classic transformer is found to be proficient at obtaining a compressed and sparse representation for the input~\cite{whiteboxtrans} to benefit the downstream tasks. 
TANGOS~\cite{TANGOS}, by regularizing neurons to focus on sparsity, also confirms this benefit.
We contend that, however, it is less capable of mining meaningful feature interactions through arithmetic operations. The importance of such interaction has been verified in various domains, \eg the serum triiodothyronine to thyroxine (T3/T4) ratio for thyroid disorder diagnosis~\cite{thyroid} and the body mass index (BMI) for obesity assessment. In summary, tabular deep learning with the classic transformer is somehow similar to the sample space division based on raw features, and incorporating arithmetic feature interaction explicitly allows for better separation of samples in an extended space with automatically engineered features.

To validate our hypothesis, we create a synthetic dataset based on a mild feature interaction assumption inspired by~\cite{sian2022}. 
The dataset consists of eight features and responses are formulated as an additive mixture of arithmetic feature combinations that remain sparse, limited in interaction order, and deterministic.
We compare the performance of  XGBoost~\cite{xgboost}, the classic transformer, and our modified \model, a transformer-like architecture enabling arithmetic feature interaction, on this data. 
%Specifically, we divide instances into equally-sized classes according to the response values and test the classification accuracy of different approaches. 
Our results reveal that, in the presence of feature interaction, \model significantly outperforms (up to {+57\%}) the other models for fine-grained tabular data modeling. Additionally, by explicitly learning arithmetic interaction, \model also obtains substantial improvements in terms of data efficiency in training (up to {+16\%}) and generalization (up to {+20\%}) compared to its counterparts. 
We describe the details of dataset construction and our empirical results in Section~\ref{sec:syn}. These findings have clearly demonstrated the effectiveness of our \model as a general module for deep tabular learning.

The above strengths of \model are rooted in two key features that address the primary challenges posed by feature heterogeneity during model fitting. The first challenge is the risk of underfitting caused by missing essential features, while the second  is the risk of overfitting caused by irrelevant correlations in redundant features. 
To compensate for features that require arithmetic feature interaction,  we equip \model with parallel attention operators responsible for extracting meaningful additive and multiplicative interaction candidates. Along the candidate dimension, these candidates are then concatenated and fused using a down-sampling linear layer, allowing each layer of \model to capture arithmetic feature interaction effectively.
To prevent overfitting caused by feature redundancy, we drop self-attention and use two sets of prompt vectors as addition and multiplication queries. This approach gives \model constrained freedom for feature interaction and, as a side effect, optimizes both memory footprint and training efficiency. By integrating these two designs with the transformer, the resulting model could better analyze tabular data based on more accurate sample separation.

We further evaluate \model by comparison with six baseline approaches on four real-world tabular datasets,. Through our extensive experiments, we find that \model is generally effective for deep tabular learning: it could be plugged into existing transformer-based methods, such as AutoInt~\cite{autoint} and FT-Transformer~\cite{fttrans}, consistently providing improvement across all datasets. Furthermore, the two \model-enhanced approaches also consistently outperform XGBoost, which is not the case for the original backbone models. Our ablation study also confirmed the rationale of each building block of AMFormer. Finally, we demonstrate that our prompt optimization can improve training efficiency by an order of magnitude, making our approach more scalable for real-world cases.
Collectively, we believe that \model has identified a good inductive bias of deep tabular models. The main contributions of our work are as follows:
\begin{itemize}
    \item We empirically verify on synthetic data that arithmetic feature interaction is necessary for deep tabular learning from the perspectives of fine-grained data modeling, data efficiency in training, and generalization. 
    \item We implement the idea in \model, which enhances the transformer architecture with arithmetic feature interaction through the parallel additive and multiplicative attention operators and prompt-based optimization. 
    \item We also verify the effectiveness and efficiency of \model through extensive tests on real-world data.
\end{itemize}

\begin{figure*}[t]
    \centering
    % \hspace{-3ex}
    \begin{subfigure}[b]{0.3\textwidth}
         \centering
         \includegraphics[width=\textwidth]{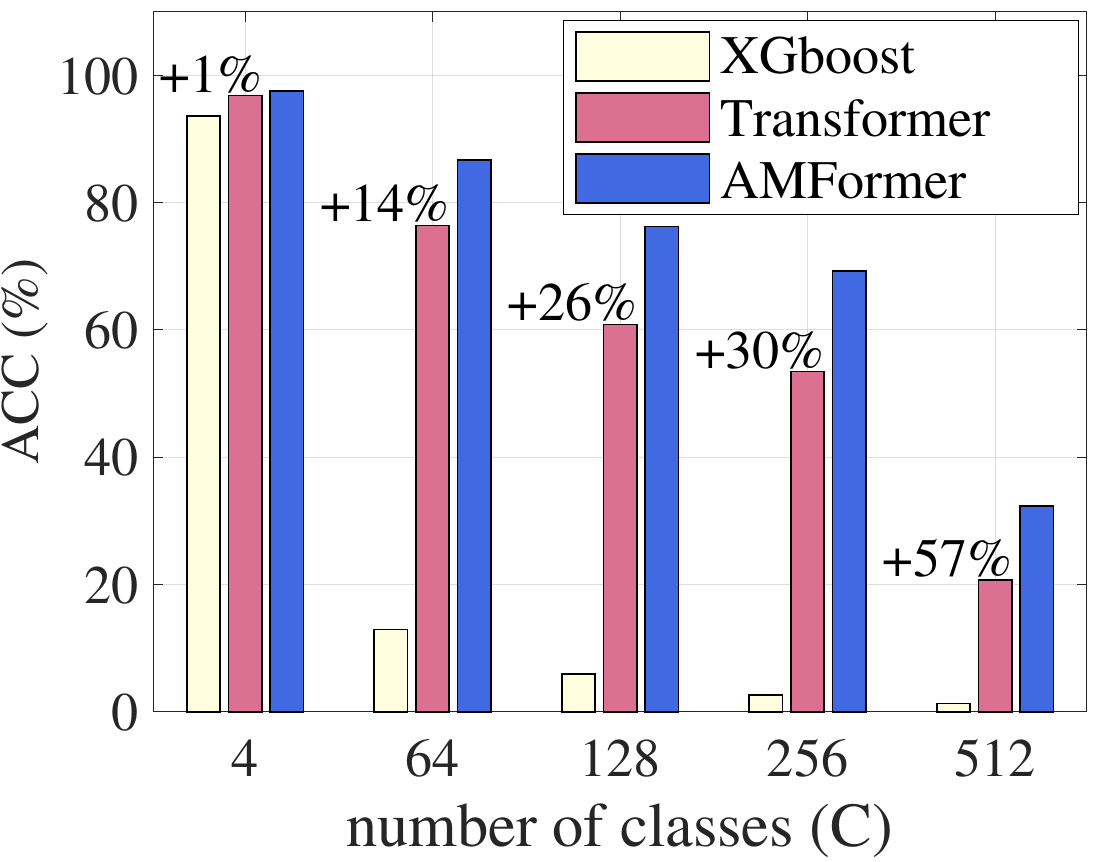}
         \caption{Fine-grained tabular data modeling}
         \label{fig:syn_finegrain}
     \end{subfigure}
     \hfill
     % \hspace{-3.75ex}
     \begin{subfigure}[b]{0.3\textwidth}
         \centering
         \includegraphics[width=\textwidth]{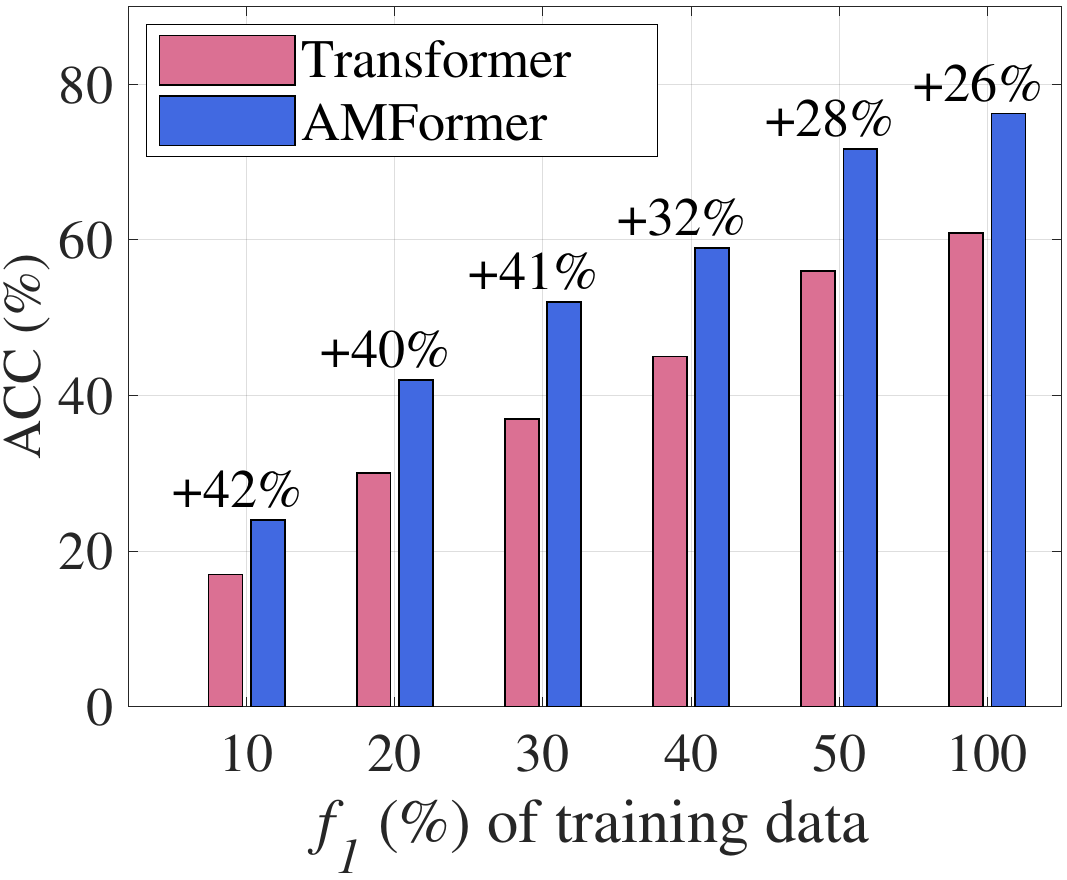}
         \caption{Data efficiency in training}
         \label{fig:syn_dataeff}
     \end{subfigure}
     \hfill
     % \hspace{-3.75ex}
     \begin{subfigure}[b]{0.3\textwidth}
         \centering
         \includegraphics[width=\textwidth]{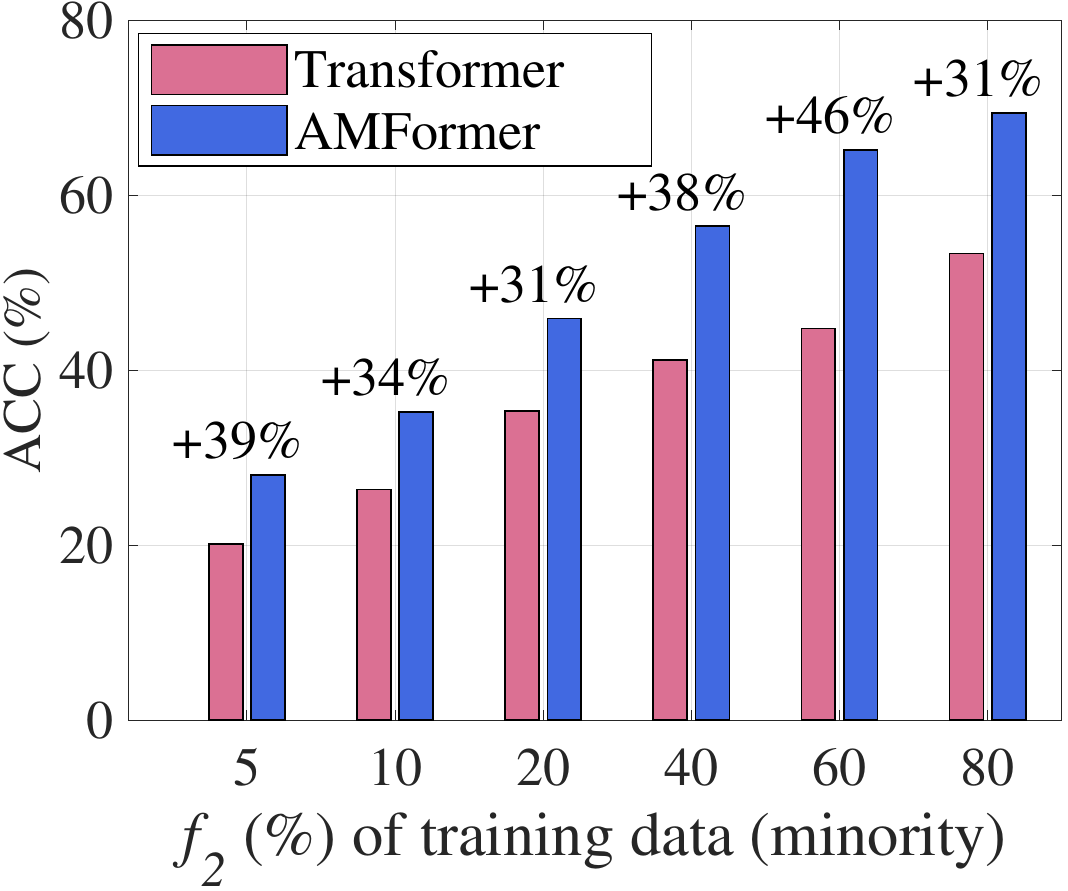}
         \caption{Generalization}
         \label{fig:syn_generalization}
     \end{subfigure}
    \caption{Results on synthetic data. The $+x\%$ in the figure are the relative improvement of \model over Transformer.}
    \label{fig:robustness}
\end{figure*}

\section{Related Work}

In this section, we review related machine-learning methods for tabular data analysis and briefly introduce the ideas of local attention that inspire our prompt-based optimization.

%on well-known methods for tabular learning as we mentioned earlier in section \ref{sec:intro}, \eg traditional and deep learning methods, and briefly overview the ideas of local attention that inspire our work.

% As we mentioned earlier in section \ref{sec:intro}, existing tabular methods can be classified into traditional methods, and deep learning models.
% % DNN-, GAM-, Tree- and transformer-based methods. DNN-based methods automatically construct high-order feature through serially connected neural networks.

\stitle{Traditional methods}. 
Tabular data could be naturally viewed as multi-dimensional vectors. Therefore, many classic machine-learning methods are applicable for mining tabular data, \eg logistic regression, decision trees, and support vector machines~\cite{prmlbook}. Since features in tabular data are of varying importance typically, generalized additive models (GAM)~\cite{GAM,ebm_kdd12} remain popular for tabular data analysis. These models are more accurate than simple linear models with the introduction of shape functions and could produce the importance of individual features as model interpretation. Pairwise interactions have also been incorporated in GAMs for better model fitness~\cite{eb2m_kdd13}. Finally, gradient-boosted tree ensemble models, such as XGBoost~\cite{xgboost}, LightGBM~\cite{lightgbm}, and CatBoost~\cite{catboost}, are usually among the most effective in this category and have been widely deployed in real-life systems.

%Tree models~\cite{xgboost, catboost} ensemble several weak classifiers and the next one continues training on the subspaces partitioned by the previous one, which provides stable and efficient performance. 

% Differently, GAM-based~\cite{GAM} methods model each features independently and finally conclude by integrating the results of each feature. Separate modeling decreases the degree of feature interaction and prevents overfitting.

\stitle{Deep learning models}.
As mentioned earlier, deep learning has been explored for dealing with tabular data recently, and the attempts could be classified into four classes.
(C1) In Wide\&Deep~\cite{widendeep} and deep factorization machines~\cite{deepfm, xdeepfm}, multi-layer perceptrons (MLPs) are stacked aside the traditional shallow components to capture high-order feature interaction. %but imbalanced ratio between high-order and low-order items can lead to overfitting.
(C2) NAM~\cite{nam2021}, NBM~\cite{nbm2022}, and SIAN~\cite{sian2022} combine GAMs with deep learning to enhance model expressive while still retraining interpretation.
(C3) Alternatively, NODE~\cite{node} and Net-DNF~\cite{Net-DNF} emulate key elements of tree models using neural networks, \eg differentiable oblivious decision trees, and disjunctive normal forms with soft neural AND/OR gates. These tree-like models could then enjoy the merits of deep learning.
(C4) Finally, relying on the global crossing capability, the transformer architecture has also been adapted to deep tabular learning. Typical models include FT-Transformer~\cite{fttrans}, AutoInt~\cite{autoint}, and DCAP~\cite{dcap}, where the attention mechanism is utilized for additive feature interaction.

% FT-Transformer~\cite{fttrans} follows ViT~\cite{vit} and utilizes class token for feature concentration within the attention module. Building upon the result from attention, AutoInt~\cite{autoint} additionally ensembles the prediction results from raw data and input embeddings. DCAP~\cite{dcap} treats the output of attention as a weight map, which is then multiplied with the input embeddings to construct new features. 
% Finally, these products are used for the final prediction.

Despite the above efforts, it still remains open for effective inductive bias of deep tabular learning. In this paper, we also consider transformer architecture as a promising candidate and further argue for the essential role of arithmetic feature interaction for deep tabular learning. Our proposed \model enhances the classic transformer with such interaction through the parallel additive and multiplication attention operators and the interaction fusion layer. This is among the first in the literature and \model has demonstrated consistently better performance compared to other transformer-based methods.

% Both MLP and transformer only construct additive feature relationship through weighed sum. Distinctly, we introduce multiplication attention to address computational restriction of vanilla attention, \ie addition only.

% we combine a multiplication attention with the original component construction in parallel to establish an arithmetic relationship.

% Local interaction limits the interaction range for tokens.
\stitle{Local attention.}
The square nature of transformer would make it less efficient for a large context, which is the case for tabular data consisting of a large number of features. 
In this situation, an efficient transformer is required.
Since words and pixels inherently possess continuity, it is natural to use local attention techniques to optimize efficiency. For instance, PVT~\cite{pvt} utilizes convolutional layers after each attention block to gather local information and gradually condenses the features. LongFormer~\cite{LongFormer} applies a moving window and calculates attention weights within the window. %, and adds global attention for some specific tokens to compensate for global interaction. 
Sparse attention~\cite{sparse-attention} introduces sparse factorizations of the full attention matrix.
%fixed sparse attention to calculate similarity for tokens within each block and ensures local tight correlation.
%
Different from the above, in this paper, we borrow the notion of prompts to localize the receptive field of features, and the number of prompts is independent of the number of features.

\section{Empirical Evaluation on Synthetic Data}\label{sec:syn}

Recall that due to feature heterogeneity, the raw feature space of tabular data might not necessarily contain all determining features. Accordingly, we argue that arithmetic feature interaction is essential for deep tabular learning, aiming at supplementing these missing features with arithmetic as the prior knowledge in a reserved stage. To validate our hypothesis, we create a synthetic dataset inspired by the assumption of a generalized additive model with multi-order while sparse feature interaction~\cite{sian2022}.

\textbf{Data construction}. 
Our constructed dataset consists of eight features and the responses $r$ are formulated as an additive mixture of arithmetic feature combinations: 
\begin{equation}
    r = \sum_{i=1}^{K} \alpha_i \cdot \prod_{j=1}^8 x_{j}^{\beta_{ij}}.
\end{equation}
We fix the number $K$ of additive terms to be much less than the number of possible feature combinations (\eg $K=5$ in this work) and sample $\alpha_i$ from a uniform distribution $U(-1,1)$. Each exponent $\beta_{ij}$ is uniformly sampled from $\{1, 2, 3, 4\}$ with a 50\% chance and is set to 0 otherwise. Thus, the expected number of involved features in each term is 4. After selecting all $\alpha$ and $\beta$ values, we randomly generate 200K instances of $(x_1, \dots, x_8, r)$ by sampling $x_j$ from a log uniform distribution between 0.5 and 2. The above setups ensure that our synthetic data remains deterministic and sparse in terms of feature interaction, adhering to a mild assumption for feature interaction. 
Finally, all instances are divided into $C$ equally-sized classes according to the response values $r$, and we further split the data into 80\%-20\% for training and testing, respectively. 

\eat{
\begin{itemize}
    \item $X_i\in{\rm LogUniform}\{0.5, 2\}$
    \item $wi\in\{0,1,2,3,4\}$
    \item $\alpha_j\in {\rm Uniform}\{-1,1\}$ 
    \item $Cls\in \{4, 64, 128, 256, 512\}$ 
\end{itemize}
}

\textbf{Results}. 
To demonstrate the necessity of arithmetic feature interaction, we compare our \model (we leave its technical details in the next section) with XGBoost and the  classic transformer on the constructed data. We first evaluate the ability of fine-grained tabular data modeling of these approaches by varying the number $C$ of classes from 4 to 512 and computing the test classification accuracy (Acc). The results are reported in  Fig.~\ref{fig:syn_finegrain}. 
When varying $C$, the Acc of all approaches decreases with the increment of $C$. The performance of XGBoost soon drops to a low level with $C=64$. This is because XGBoost utilizes raw feature values $x_j$ only, which could not handle the interaction between features and the response. Comparatively, both Transformer and \model maintain relatively high Acc with larger $C$, owing to the automatic feature engineering capacity of neural networks. We also find that \model is consistently better than Transformer, with larger improvement for higher $C$. This indicates the essential role of arithmetic feature interaction for fine-grained tabular data modeling.

\begin{figure*}[t]
    \centering
    \includegraphics[width=0.9\textwidth]{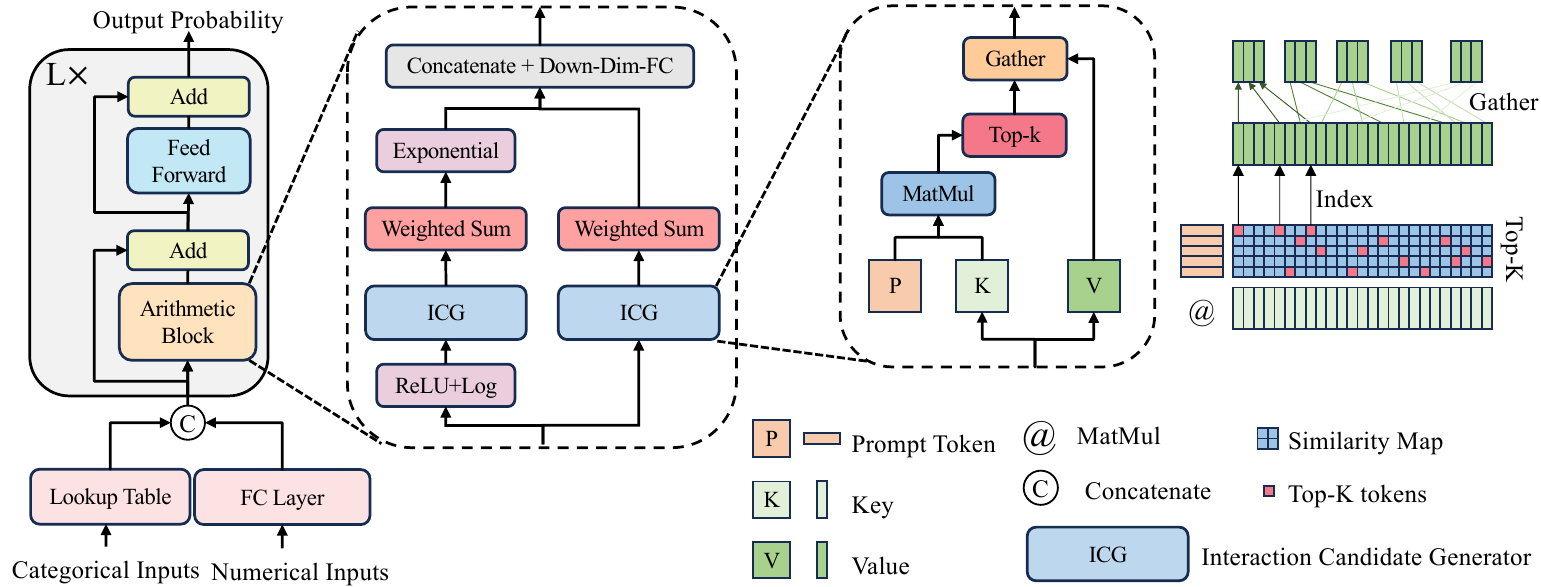}
    \caption{The overview of \model. $L$ is the layer number. 
    }
    \label{fig:overview}
\end{figure*}

In addition, through learning meaningful interaction patterns, we also expect \model to have better training data efficiency and generalization to minority classes. 
To verify data efficiency, we fix {$C=128$}, vary the fraction $f_1$ of training data from 10\% to 50\%, and compute the test Acc of Transformer and \model, reported in  Fig.~\ref{fig:syn_dataeff}. We omit XGBoost due to its uncompetitive performance. When varying $f_1$, the Acc of both approaches increases with the increment of $f_1$ as expected. Moreover, we observe consistently better relative improvement with less fraction of training data, \eg +40\% with $f_1 \le 30\%$ vs. the baseline +26\%. Indeed, \model trained with 40\% data is almost on par with Transformer on all data.

To verify generalization, we also fix {$C=128$}, manually turn half of the classes to minority classes by only reserving a fraction $f_2$ of training data, and compute the test Acc of Transformer and \model on minority classes, reported in Fig.~\ref{fig:syn_generalization}. Note that training data on rest classes and test data remain unfiltered. Similarly, we observe consistently better performance, in both absolute Acc or relative improvement, by \model for minority classes.

From the above empirical evaluation, we believe that explicitly integrating arithmetic feature interaction is necessary for deep tabular learning.

\section{Methodology}\label{sec:method}

We now present the technical details of our \model. The framework overview is given in Fig.~\ref{fig:overview}, which closely resembles the classic Transformer architecture except for the Arithmetic Block. With $d$ denoting the dimensionality of hidden states, \model initiates the process by transforming raw features into representative embeddings, \ie applying a 1-in-$d$-out linear layer for numerical features and a $d$-dimensional embedding lookup table for categorical features. Subsequently, these initial embeddings are processed through $L$ sequential layers, which serve to augment them with vital context and interactive elements. Within each of these layers, an arithmetic block that executes parallel additive and multiplicative attention is adopted to deliberately foster arithmetic feature interactions. Residual connections and feed-forward networks are reserved to facilitate the flow of gradients and augment feature representation. Finally, \model employs either a classification or a regression head to generate the final output based on these enriched embeddings.
The key components in the arithmetic block include parallel attention and prompt tokens, which we will further discuss in the following.

\subsection{Parallel Attention}\label{sec:architecture}

The parallel attention is responsible for facilitating arithmetic feature interaction in \model.
The fundamental concept involves harnessing two parallel attention streams, each dedicated to computing either additive or multiplicative interaction candidates based on input feature embeddings. These computed candidates are subsequently concatenated and undergo further integration through a fully-connected (FC) layer. Consequently, the outputs of the FC layer can represent diverse combinations of input features achieved through arithmetic operations.

Formally, let $N$ denote the number of input features in a specific layer $l \in \{1, \dots, L\}$ and $X \in \mathbb{R}^{N \times d}$ the corresponding feature embedding matrix. As the layers in \model are structurally identical, differing only in their parameters, we forego mentioning the layer index for ease of notation. 
Recall that the classic attention~\cite{Attention} is inherently additive. We therefore derive log-scaled embeddings in the multiplicative stream~\cite{nalu2018}:
\begin{equation}
    X_{log} = {\rm log}({\rm ReLU}(X)+\epsilon),
    \label{eq:log}
\end{equation}
where $\epsilon$ prevents $\log 0$. Classic attention in the log space, combined with an exponential operator, is then capable of learning multiplicative interaction.

We next elaborate on the shared attention part of the two streams, which aims to prepare useful additive and multiplicative feature interaction for the prediction task. Taking the additive stream for instance, we first generate the query $Q=XW^Q$, key $K=XW^K$, and value $V=XW^V$ embeddings with input embeddings $X$ and trainable parameters $W^{Q/K/V} \in \mathbb{R}^{d \times d}$. The product $QK^T$ gives the affinity between input features and could be understood as the likelihood of yielding meaningful interaction in our case. The output by the additive stream, calculated as the weighted sum of value embeddings according to the normalized product:
\begin{equation}
    O^A = {\rm softmax}(\frac{QK^T}{\sqrt{d}})V \in \mathbb{R}^{N \times d},
    % O^A = {\rm softmax}({QK^T}\div{\sqrt{d}})V \in \mathbb{R}^{N \times d},
    \label{eq:attention}
\end{equation}
enriches each input feature with vital interactive information from other features and, hence, could be regarded as additive interaction candidates.

The above soft attention establishes a connection between every pair of features, while the interaction on tabular data is mostly sparse~\cite{nam2021,nbm2022}. Inspired, we further revise Eq.~(\ref{eq:attention}) into hard attention. That is, we retrieve and retain the top-$k$ highest entries in each row of $QK^T$ while masking other entries with a large negative constant, \ie becoming 0 after softmax. As such, each feature could only interact with $k$ features and it suffices to choose a small integer for hyperparameter $k$, being independent with the number $N$ of input features, to ensure sparse interaction.

Similarly, we could obtain the multiplicative stream output $O^M \in \mathbb{R}^{N \times d}$ by using the log-scaled $X_{log}$, another set of trainable parameters, and an additional exponential operator.
The central idea of our \model is to provide complete arithmetic ability in every single layer. To achieve the goal, we further concatenate the two outputs $O^A$ and $O^M$ and apply an FC layer along the candidate dimension:
\begin{equation}
    O = {\rm FC}({\rm VConcat}([O^A, O^M])^T)^T.
    \label{eq:combine}
\end{equation}
Note that ${\rm VConcat}([O^A, O^M])$ performs vertical concatenation, resulting in a shape of ${2N \times d}$, ${\rm FC}$ is a fully-connected layer that also decreases the number of dimension from $2N$ to $N$, and the two transposition operators assure the fusion of interaction candidates. 
To summarize, the parallel attention in the arithmetic block first extracts additive and multiplicative interaction candidates independently, and the candidates are subsequently mixed to facilitate complete arithmetic feature interaction in a single layer of \model.

\subsection{Optimization with Prompt Tokens}\label{sec:prompt}

The architecture described by far potentially faces a couple of practical challenges. Initially, the self-attention mechanism has a complexity that grows quadratically with the number of features, specifically $O(N^2d)$, resulting in inefficiencies in time and memory for large feature sets~\cite{YangZZXJY23,YangSZFZXY23}. Additionally, tabular datasets often exhibit interaction patterns that are data-invariant, \eg those relating to domain patterns, while the current purely data-reliant architecture has trouble capturing them.
To tackle the two issues, we have incorporated an optimization strategy that utilizes prompt tokens within \model.

The main idea is to substitute the data-dependent query matrix $Q$ with a set of trainable prompt token embeddings, represented as a parameter matrix $P \in \mathbb{R}^{N_p \times d}$. Each of these prompt tokens is designed to facilitate the creation of valuable additive or multiplicative interactions among features, considering up to $k$ features at a time. The quantity $N_p$ of prompt tokens could be much less than the overall number of features on large datasets. % \footnote{In practice, we recommend fixing $N_p=N$ for datasets with a few hundred features. In cases of larger datasets, it is suggested to start with $N_p$ at 256 or 512 and then reduce this number by half for each ensuing layer.} 
Consequently, the time and memory complexities associated with \model become linear relative to the number $N$ of input features, thus enhancing the model's capability to handle extensive datasets. Additionally, this approach enables \model to inherently learn and establish patterns of feature interaction that are consistent across different samples. The strategic use of prompt tokens in conjunction with the top-k selection also allows \model to disregard immaterial correlations in the data. This not only helps in preventing overfitting but also improves the model's resilience against data noise~\cite{cheng2023robust,qian2023robust}.

In practice, we recommend fixing $N_p=N$ for datasets with a few hundred features. In cases of larger datasets, it is suggested to start with $N_p$ at 256 or 512 and then reduce this number by half for each ensuing layer.

\section{Experiments}

\begin{table*}[t]
    \centering
    \begin{tabular}{cccccccc}
        \hline
        Dataset    & Task type & Metric            &\# Train&\# Valid&\# Test &\# Num. features &\# Cate. features \\
        \hline
        EP    & binary classification   & Acc \eat{$\uparrow$} &320,000 &80,000  &100,000 & 2,000  & / \\
        HC    & binary classification   & AUC \eat{$\uparrow$} &200,496 &45,512  &61,503  & 104   & 16	\\
        CO    & multi-classification & Acc  \eat{$\uparrow$} &371,847 &92,962  &116,203 & 54    & /	\\
        MI    & regression & MSE \eat{$\downarrow$} &72,3412 &235,259 &24,1521 & 136   & /	\\
        \hline
    \end{tabular}
    \caption{Dataset statistics and evaluation settings.} 
    %Notation: ACC $\sim$ accuracy, AUC $\sim$ area under receiver operating characteristic, MSE $\sim$ mean squared error, Num $\sim$ numerical, Cate $\sim$ categorical.
    %\# Num. and \# Cate. denote the numbers of numerical and categorical features, respectively.
    \label{tab:dataset}
\end{table*}

% \begin{table*}[ht]
%     \centering
%     \begin{tabular}{clllll|ll}
%         % \toprule
%         \hline
%         Dataset     & XGBoost & FT-Trans & AutoInt & DCN-V2 & DCAP & Ours + FT-Trans & Ours + AutoInt\\
%         \hline
%         EP     &87.32     &89.05   &88.48  &88.22  &\textcolor{orange}{89.24}  &\redtextbf{89.83}, $\Delta=0.78$   &\textcolor{red}{89.71}, $\Delta=1.23$\\
%         HC        &74.59 	   &75.07 	&75.01 	&72.34	&\textcolor{red}{75.63}	&\redtextbf{75.67}, $\Delta=0.60$ 	&\textcolor{orange}{75.57}, $\Delta=0.56$\\
%         CO     &\textcolor{orange}{96.72}     &96.60 	&92.40 	&90.78 	&96.21  &\textcolor{red}{97.26}, $\Delta=0.66$   &\redtextbf{97.36}, $\Delta=4.96$\\
%         MI   &\textcolor{orange}{0.5642}    &0.5717 	&0.5864 &0.6043 &0.5753 &\redtextbf{0.5557}, $\Delta=0.0160$	&\textcolor{red}{0.5606}, $\Delta=0.0258$\\
%         \hline
%         Rank 
%         &$4.8\pm2.1$
%         &$4.0\pm0.0$
%         &$5.5\pm0.6$
%         &$6.8\pm0.5$
%         &\textcolor{orange}{$3.8\pm1.5$}
%         &\textcolor{red}{\boldmath{$1.3\pm0.5$}}
%         &\textcolor{red}{$2.0\pm0.8$}\\
%         \hline
%         % \bottomrule
%     \end{tabular}
%     \caption{Performance comparison of \model and known methods. The \redtextbf{best}, \textcolor{red}{second-best}, \textcolor{orange}{third-best} results are highlighted by bolding or coloring. $\Delta$ means the gain from previous methods.}
%     \label{tab:summary}
% \end{table*}

\begin{table*}[t]
    \centering
    \begin{tabular}{ccccccc|cc}
        % \toprule
        \hline
        Dataset   & XGBoost   & NODE      & DCN-V2    & DCAP              & AutoInt   & FT-Trans.      &  \modelauto         & \modelftt       \\
        \hline
        EP $\uparrow$   &87.32 (8)      & 89.60 (3)     &88.22 (7)      &89.24 (4)              &88.48 (6)      &89.05 (5)      &\underline{89.71} (2)      &\textbf{89.83} (1) \\
        HC $\uparrow$   &74.59 (7) 	    & 74.93 (6)     &72.34 (8)      &\underline{75.63} (2)   &75.01 (5)      &75.07 (4)	    &75.57 (3)                  &\textbf{75.67} (1)\\
        CO $\uparrow$   &96.72 (3)      & 92.31 (7)     &90.78 (8) 	    &96.21 (5)              &92.40 (6)	    &96.60 (4)      &\textbf{97.36} (1)         &\underline{97.26} (2) \\
        MI $\downarrow$ &0.5642 (3)     & 0.5644 (4)    &0.6043 (8)     &0.5753 (6)             &0.5864 (7)     &0.5717 (5)     &\underline{0.5606} (2)     &\textbf{0.5557} (1)\\ 
        \hline
        Rank $\pm$ std  &$5.3\pm2.6$    &$5.0\pm1.8$    &$7.8\pm0.5$    &$4.3\pm1.7$            &$6.0\pm0.8$    &$4.5\pm0.6$    &\underline{$2.0\pm0.8$}    &\boldmath{$1.3\pm0.5$} \\
        \hline
        % \bottomrule
    \end{tabular}
    \caption{Performance comparison of \model and known methods. The numbers inside parentheses are the ranks of performance. The \textbf{best} and \underline{second-best} results are highlighted.}
    % The best and second-best results are highlighted in bold and underlined, respectively.}
    \label{tab:summary}
\end{table*}

%In section \ref{sec:syn}, we have validated that vanilla self-attention is insufficient to handle production relationships, and set up additional experiments to verify the robustness and generalization of our approach in handling multiplication.
% 
In this section, we further evaluate our \model in real-world scenarios.
% containing binary classification, multi-classification and regression, with varying numbers of columns.
Using four datasets, we conduct extensive experiments to evaluate: (i) the overall effectiveness of \model compared with other competitive methods, (ii) the rationale of each building component in \model, (iii) the impacts of the prompt-based optimization in effectiveness and scalability, and (iv) the sensitivity to key parameters.
% Three sets of experiments are conducted to evaluate: (i) production ability of vanilla attention and our attention, (ii) the overall effectiveness of \model with well-known state-of-the-art methods, (iii) computational efficiency with transformer-based methods on the large amount number of tokens dataset.

\subsection{Experimental Setup}
\stitle{Datasets.} We choose four real-world datasets for evaluation.
\begin{itemize}
    \item {\bf EP} (Epsilon) is a simulated physics dataset from~\cite{epsilon_url}, which uses 2,000 normalized numerical features for binary classification. %to predict binary simulated physics experiment results.
    \item {\bf HC} (Home Credit Default Risk) uses both numerical and categorical features to predict clients' binary repayment abilities, with a ratio of around 1:10 for positive to negative samples~\cite{hcdr}.
    \item {\bf CO} (Covtype) is a multi-classification dataset that utilizes surrounding characteristics to predict types of trees growing in an area~\cite{covtype}.
    \item {\bf MI} (MSLR-WEB10K) is a learn-to-rank dataset for query-URL relevance ranking~\cite{microsoft}. It contains 136 numerical features and relevance scores are drawn from $\{0,1,2,3,4\}$.  
\end{itemize}

We adopt the same learning tasks, \ie binary/multi-classification and regression, and metrics, \ie accuracy (Acc), area under the ROC curve (AUC), and mean square error (MSE), as previous studies. Table~\ref{tab:dataset} summarizes the statistics and evaluation settings of these datasets. 
We normalize the numerical features to have zero mean and unit variance before feeding into models.

% \footnote{\textcolor{red}{Differnet metrics for classification task due to the differing balance of positive and negative examples}}

\stitle{Baselines}. 
We compare \model to a variety of baseline methods, including the competitive tree ensemble method XGBoost~\cite{xgboost}, differentiable tree model NODE~\cite{node}, transformer-based approaches  AutoInt~\cite{autoint} and FT-Transformer~\cite{fttrans}, and the recent deep cross nets DCN-V2~\cite{dcnv2} and DCAP~\cite{dcap}. %  based on MLP and transformer, respectively
Our \model is a general deep tabular learning module and we plug it into FT-Transformer and AutoInt, leading to two variants \modelauto and \modelftt for comparison, respectively.

% FT-Transformer~\cite{fttrans} follows ViT~\cite{vit} and utilizes class token for feature concentration within the attention module. Building upon the result from attention, AutoInt~\cite{autoint} additionally ensembles the prediction results from raw data and input embeddings. DCAP~\cite{dcap} treats the output of attention as a weight map, which is then multiplied with the input embeddings to construct new features. 

\stitle{Implementation.}
All tested models are implemented with PyTorch v1.12~\cite{pytorch}. 
%Our AM-Former is available at GitHub\footnote{\url{https://github.com/aigc-apps/AMFormer}}.
We use the recommended model parameters in the original papers for baseline methods and those of XGBoost follow~\cite{fttrans}. Notably, AutoInt and FT-Transformer use a dimensionality of 32 and 192 for embeddings, respectively, which are inherited in \modelauto and \modelftt. 
We adopt Adam with betas=(0.9, 0.999) and eps=1e{-8} for optimization. The learning rate first linearly increases to 1e-3 in the first 1k steps and then decays by 90\% every 20k steps by default, except for the HC data with an initial 1e-4 and 4k decaying steps.  
The default batch size is 512, which reduces to 32 for transformer-based methods on the EP data due to GPU-memory limitation.
We report the detailed hyper-parameters of all methods in the supplement.
All tests are conducted on a machine with 104 Intel(R) Xeon(R) Platinum 8269CY CPUs and an NVIDIA Tesla A100-SXM-40GB.  

% Embedding dimension and dimension in attention are 32 for AutoInt (following the setup described in the paper) and 192 for FT-Transformer and DCAP, 8 heads are used in multi-head attention. 
% The learning rate is $\rm 1e^{-4}$ for HCDR and $\rm 1e-3$ otherwise and is decayed to 10\% after 4k iterations for HCDR dataset and 20k iterations otherwise, and the max iterations are 10k for HCDR and 40k for others.
% Layer numbers are 3 for all transformer-based models for real world datasets and 6 for synthetic dataset. Dropout rates for FFN and attention are 0.1 and 0.2, respectively. 

We next present our findings.

% \stitle{Baseline.} 
% In order to test the performance on different tasks and the effectiveness of dealing large-feature data, we choose four data with binary classification, multi classification, regression tasks and the feature number varies from 54 to 2000. Table \ref{tab:dataset} shows the four datasets used in the experiments. The metrics of each dataset follow the previous works. All the presented results are from the test set.

% \begin{table}[t]
%     \centering
%     \begin{tabular}{cccc}
%         % \toprule
%         \hline
%         Method     & Training(hr) & MFLOPS & RAM(GB) \\
%         \hline
%         AutoInt(A)                     &16.81	         &	203.65			&27.8\\
%         FT-Trans(F)                    &19.64	   &	173.65		&	36.5\\
%         DCAP                        &15.76			&	100.61		&	28.5\\
%         \hline
%         Ours-A              & 1.87	&26.14	&7.5\\
%         Ours-F             & 1.72	&22.04	&7.5\\
%         \hline
%         % Ours-A($\Delta\%$)  & 11.12	    &12.83      &27.2\\
%         % Ours-F($\Delta\%$) & 8.76      &12.69      &20.6\\
%         % \hline
%         $\Delta$-A  & 11.12\%	    &12.83\%      &27.2\%\\
%         $\Delta$-F  & 8.76\%      &12.69\%      &20.6\%\\
%         \hline
%         % \bottomrule
%     \end{tabular}
%     \caption{Computational cost and training time on Epsilon dataset (2000 tokens). $\Delta$ indicates the ratio between model with our \model and original model.}
%     \label{tab:flop}
% \end{table}

\subsection{Comparison with Known Methods}

In the first set of tests, we evaluate the overall effectiveness of our approach by comparing its variants with the considered baselines. %(including state-of-the-art ones).
The metrics on the test set are presented in Table~\ref{tab:summary} and we conclude the following. %in which Ours(F) and Ours(A) denote two variants of our approach based on the FT-Transformer and AutoInt, respectively. Even though DCAP is a transformer-based method, the inner attention module is used to generate weight map, whose output is then directly multiplied with the input. Therefore, we do not transfer our \model onto it. 

First, DCN-V2 performs the worst compared with other tested approaches. Note that it utilizes MLP to capture high-order feature interaction, which turns out to be less effective for tabular data. AutoInt only preserves the multi-head self-attention while dropping the feed-forward net and residual connection from its transformer architecture, and its effectiveness remains not competitive either. Inspired, we keep these operators in our \model.

For the rest baselines, we observe inconsistent performance across different datasets. The top-performing methods are diverse, \eg XGBoost on CO and MI, NODE on EP, and DCAP on HC. This result has somehow demonstrated the challenge of identifying a unified modeling bias for tabular data analysis.
Overall, we find that DCAP and FT-Transformer are the best among these baselines according to the average performance rank on all datasets. Both approaches are based on the transformer architecture, indicating its potential for deep tabular learning. 
However, it is worth noting that none of the deep models could outperform XGBoost on all datasets.

Finally, the two variants of \model are consistently better than the six tested baselines on the four datasets, except for \modelauto on the HC data where it slightly underperforms compared to DCAP. 
This implies that \model consistently enhances the performance of the two backbone models on all datasets. 
Specifically, in classification tasks, \model improves the accuracy or AUC of AutoInt and FT-transformer by at least 0.5\%,  with improvements of up to 1.23\% and 4.96\% observed on the EP and CO data for AutoInt. In regression tasks, the MSE of the two backbone models decreases by more than 0.016. The effectiveness and stability of \model result in significantly better performance rankings for \modelauto and \modelftt compared to all existing approaches, marking a significant milestone for deep learning on tabular data. From our perspective, we believe that \model effectively addresses the question of whether deep learning is necessary for tabular data.

\subsection{Ablation Study}

\begin{table}[t]
    \centering
    \begin{tabular}{cccccc}
        % \toprule
        \hline
        Backbone & Add. & Multi. & Prompt & EP $\uparrow$ & MI $\downarrow$ \\
        \hline
        \multirow{6}{*}{AutoInt} 
            &   \checkmark  &   -           &   -           &    88.48              & 0.5864\\
            &   \checkmark  &   -           &   \checkmark  &   89.61             & 0.5638 \\
            &   -           &   \checkmark  &   -           &   89.53              & 0.5748 \\
            &   -           & \checkmark    &   \checkmark  &   \underline{89.65}  & \underline{0.5631}\\
            &   \checkmark  & \checkmark    &   -           &   89.55             & 0.5639  \\
            &   \checkmark  & \checkmark    &   \checkmark  &   \textbf{89.71}     & \textbf{0.5606}\\
        \hline
        \multirow{6}{*}{FT-Trans.} 
            &   \checkmark  &   -           &   -           &   89.05              & 0.5717\\
            &   \checkmark  &   -           &   \checkmark  &   \underline{89.80}  & 0.5633 \\
            &   -           &   \checkmark  &   -           &   50.05              & 0.5624\\
            &   -           & \checkmark    &   \checkmark  &   50.05              & 0.5605\\
            &   \checkmark  & \checkmark    &   -           &   89.58              & \underline{0.5585}\\
            &   \checkmark  & \checkmark    &   \checkmark  &  \textbf{89.83}      &  \textbf{0.5557}\\
        \hline
    \end{tabular}
    \caption{Results of ablation study on EP and MI. 
    % The \redtextbf{best} and \textcolor{red}{second-best} are highlighted.
    }
    \label{tab:ablation}
\end{table}

We next conduct an ablation study to evaluate the impacts of the building component within our \model, \ie the additive attention, the multiplicative attention, and the prompt-based optimization. Similarly, we consider two backbone models and evaluate the test metrics by using different combinations of the components on the EP and MI datasets. Note that EP has the largest number of features and MI is the only regression dataset.
The results are reported in Table~\ref{tab:ablation} and we find the following.

First, we find that multiplicative interaction is important, and using multiplicative attention alone could obtain better results than using classic additive attention alone in three of the four testing cases (row 1 vs. row 3), except for using FT-Transformer as the base model on EP. %\marked{We guess that this is possibly due to xxx.}
Moreover, using parallel additive and multiplicative attention consistently improves the effectiveness in all scenarios (row 5 vs. 1\&3), indicating the usefulness of arithmetic feature interaction for tabular data analysis.
In addition, our prompt-based optimization also leads to consistent performance improvement (rows 1\&3\&5 vs. 2\&4\&6, respectively). This is because prompts could stabilize the interaction patterns which do not vary with examples and this is very important for tabular data. Moreover, when dealing with a large number of features, a small number of prompts could reduce irrelevant interactions in redundant features.

\begin{table}[t]
    \centering
    \begin{tabular}{lclll}
        % \toprule
        \hline
        Method      & Prompt & Train (hr) & GFLOPS & GPU-M\\
        \hline
        AutoInt     & -                 & 16.81	    & 7.2       & 33 GB    \\
        FT-Trans    & -                 & 19.64	    & 9.7       & 34 GB\\
        \modelauto       & -                 & 25.77     & 12.3      & 36 GB\\
        \modelftt       & -                 & 20.88     & 11.3      & 38 GB\\
        \hline
        \modelauto     & \checkmark        & 2.93      & 1.3       & 17 GB\\
        \modelftt       & \checkmark        & 3.11      & 0.7       & 18 GB\\
        \hline
        \modelauto\%  & \checkmark                  & 11.37\%   &10.45\%    & 51.51\%   \\
        \modelftt\%  & \checkmark                  & 14.89\%   &6.16\%     & 52.94\%\\
        \hline
        % \bottomrule
    \end{tabular}
    \caption{Scalability evaluation of prompt-based optimization
    %$\Delta$ indicates the ratio between \model with prompt and \model without prompt.
    }
    \label{tab:flop}
\end{table}

% \vspace{-2.75ex}
% \subsection{Effectiveness on Large Feature Dataset}
% \vspace{-1}
% \subsection{Efficiency Impact of Prompt}
%\stitle{Efficiency impact of prompt.}
The complexity of attention has always been a very efficiency-critical part. With an $\mathcal{O}(N^{2})$ complexity to the number $N$ of tokens/features, the computational cost will increase quadratically with the increase of token number, and the attention map will also occupy a lot of GPU memory during training. 
From Table~\ref{tab:flop} we find that the overall training time for FT-Transformer and AutoInt exceeds 15 hours. For our \model without prompts, the training time even exceeds 20 hours. Our \model incorporates the additional multiplicative attention, which requires more FLOPs and results in longer training time. After applying the prompt optimization to limit feature interaction, our \model reduces computational resource by approximately 90\% and 94\% and achieves approximately 7 times faster training speed, \ie only around 3 hours of training time.
% resulting in a training time of around 3 hours, bringing about a speedup of approximately 7 times.
% 
Moreover, with the same batchsize, our \model needs only 17 GB GPU-memory while its counterparts take up at least 33 GB GPU-memory. Less memory usage makes it possible to train and employ our \model on less-powerful devices.

% of these three methods are beyond 15 hours while the training time has been reduced to less than two hours by applying our \model to replace the vanilla attention in FT-transformer and AutoInt. Moreover, with the same batchsize, our \model needs only 7.5GB GPU-memory while they take up at least 27GB GPU-memory. In conclusion, our \model use only 10\% computational resource and training time but can also bring a huge improvement on Epsilon dataset. Meanwhile, about 20\% GPU-memory usage make it possible to train and employ our \model on other less-powerful devices.

% \begin{table}[ht]
%     \centering
%     \label{tab:token_number}
%     \begin{tabular}{cccc}
%         % \toprule
%         \hline
%         \#Input token     & $N_c, l=1$ & $N_c, l=2$ & $N_c, l=3$ \\
%         2000                     &256	         &	128			&128\\
%         \hline
%         \multirow{2}{*}{topk(sum)} & $-$ & $\searrow$ & $\searrow$ \\
%                             &128(default)	   &	256/4=64		&	128/4=32\\
%         \hline
%         % \multirow{2}{*}{topk(prod)} & $\downarrow$ & $\downarrow$ & $\downarrow$ \\
%         %                         &128/2=64			&	64/2=32		&	32/2=16\\
%         \multirow{2}{*}{topk(prod)} & $-$ & $-$ & $-$ \\
%                                 &8			&8		&8\\
%         \hline
%     \end{tabular}
%     \caption{Default cluster number and topk settings on \model on Epsilon dataset.}
% \end{table}

\begin{figure}[tb!]
    \centering
    % \hspace{-3ex}
    \begin{subfigure}[b]{0.46\linewidth}
         \centering
         \includegraphics[width=\linewidth]{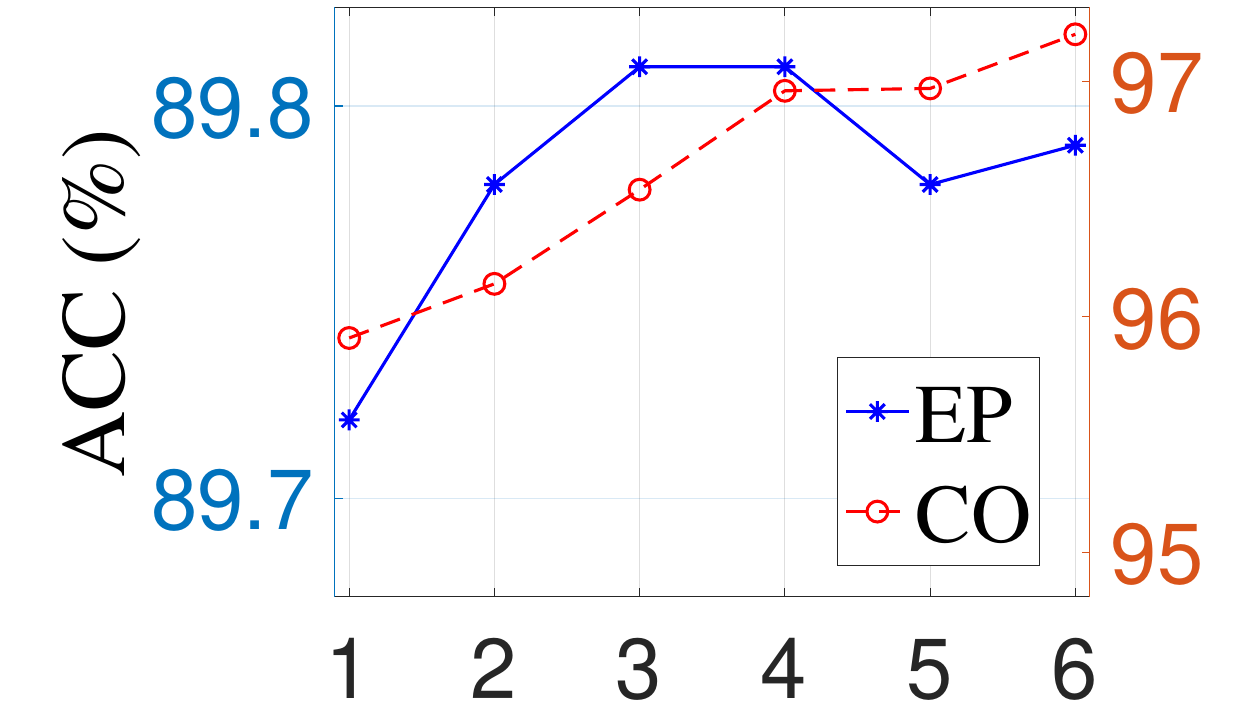}
         \caption{Layer number $L$} %  on EP \& CO
         \label{fig:layernum}
     \end{subfigure}
     \begin{subfigure}[b]{0.46\linewidth}
         \centering
         \includegraphics[width=\linewidth]{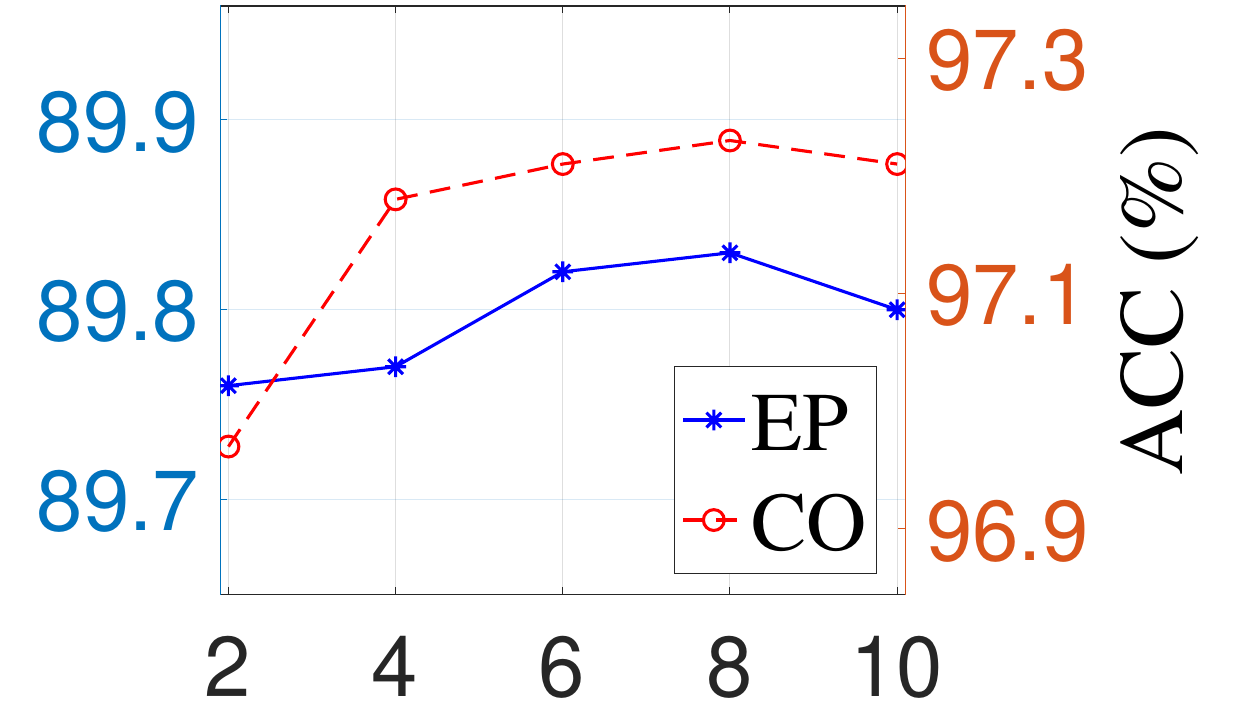}
         \caption{Parameter $k$} % on EP \& CO
         \label{fig:topk}
     \end{subfigure}
    \caption{Impacts of layer number $L$ and parameter $k$.}
    \label{fig:param sen}    
\end{figure}

\begin{table}[t]
    \centering
    \begin{tabular}{ccccccc}
        % \toprule
        \hline          
        % Token & 1024 & 512 & 256 & 128 & 64 & 32 \\
        $N_P$ & 32 & 64 & 128 & 256 & 512 & 1024 \\
        EP & 89.75 & 89.79 & 89.77 & \underline{89.81} & \textbf{89.83} & 89.76 \\
        % EP & 89.76 & 89.83 & 89.81 & 89.77 & 89.79 & 89.75 \\
        % \hline
        % Token & - & +5 & +10 & +15 & +20 & +25 \\
        % CO & 96.95 & 96.94 & 96.96 & 96.84 & 96.95 & 96.98\\
        \hline
    \end{tabular}
    \caption{Impact of the number $N_P$ of prompt tokens. %\#P indicates the prompt tokens number for the first layer.
    }
    \label{tab:token number}
\end{table}

\subsection{Parameter Sensitivity}
Finally, we evaluate the parameter sensitivity of \model. 
To examine the impact of $L$, we vary $L$ from 1 to 6, fix the other parameters to their default values, and test the Acc results on the EP and CO dataset, as shown in Fig.~\ref{fig:layernum}.
When increasing $L$, the Acc first increases and then decreases on EP while it continues increasing on CO.
The best Acc is attained at $L=3$ for EP and $L=6$ for CO. 
Note that the accuracy metric on the CO dataset starts to plateau at $L=4$. Overall, we believe that approximately $3\sim4$ layers are sufficient for feature extraction.

When dealing with large datasets, we set the number of prompt tokens as $N_P$ in the first layer, and then reduce this number by half for each subsequent layer.
As shown in Table~\ref{tab:token number}, on the large EP data, a small $N_P$ ($<256$) results in slightly worse performance caused by insufficient learning, while the performance tends to saturate around 256.
However, increasing $N_P$ further leads to a decrease in performance, because large $N_P$ tends to cause redundancy and overfitting. 
Although 512 tokens exhibit better performance than 256, the trade-off between performance and efficiency makes 256 a balanced choice.

We further present the sensitivity results of parameter $k$, \ie top-k, for forming feature interaction.
The results on EP and CO are shown in Fig.~\ref{fig:topk}, from which we find that the Acc increases first from 2 to 8 and then decreases slightly.
Overall, large $k$ is more suitable for \model. This is because the impacts of less relevant features are decreased by the affinity weights. However, there is a risk of overfitting in practical training and larger $k$ can then have a negative impact. We thus recommend $k=8$ by default.

% The results for the other hyper-parameters (\ie $\lambda$, $\alpha$, and $\beta$) are given in the Supplementary Material. 

% \begin{figure}[t]
%     \centering
%     \includegraphics[width=\linewidth]{LaTeX/images/layer_num.pdf}
%     \caption{ 
%     % Orange and blue rectangles on the right side are inputs and prompt tokens.
%     }
%     \label{fig:layer_num}
% \end{figure}

% \begin{table}[ht]
%     \centering
%     % \label{tab:topk}
%     \begin{tabular}{ccccccc}
%         % \toprule
%         \hline          
%         Dataset & k=2& k=4 & k=6 & k=8 & k=10\\
%         \hline
%         EP &  \\
%         CO & 97.18 &   \\
%         \hline
%     \end{tabular}
%     \caption{}
% \end{table}

% \begin{figure}[t]
%     \centering
%     % \hspace{-3ex}
%     \begin{subfigure}[b]{0.45\linewidth}
%          \centering
%          \includegraphics[width=\linewidth]{LaTeX/images/cov_less_train.pdf}
%          \caption{Full size.}
%          \label{fig:cov less train}
%      \end{subfigure}
%      %\hfill
%      % \hspace{-3.75ex}
%      \begin{subfigure}[b]{0.45\linewidth}
%          \centering
%          \includegraphics[width=\linewidth]{LaTeX/images/less_train.pdf}
%          \caption{Less train set.}
%          \label{fig:syn less train}
%      \end{subfigure}
     
%     \caption{Results on Synthetic dataset. Dashed lines in Fig.\ref{fig:syn minority abs} indicate the accuracy with full size training.}
%     \label{fig:robustness}
% \end{figure}

\section{Conclusion}
This paper studied the effective inductive bias of deep models on tabular data. We hypothesized that arithmetic feature interaction is necessary for tabular deep learning and integrated this idea in the transformer architecture to derive \model. We verified the effectiveness of \model on both synthetic and real-world data. The results of our synthetic data demonstrated its better capacity for fine-grained tabular data modeling, data efficiency in training, and generalization.  Moreover, extensive experiments on real-world data further confirmed its consistent effectiveness, the rationale behind each building block, and the scalability to handle large-scale data. We thus believe that \model has established a strong inductive bias for deep tabular learning.

%We first construct a synthetic dataset to validate our hypothesis that Transformer is unable to deal with multiplication interaction, which is significant in real-world scene. To deal with it, we propose our \model with para

\section*{Acknowledgments}

This research was partially supported by National Natural Science Foundation of China under grants No. 62106218 and No. 62132017, Zhejiang Key R\&D Program of China under grant No. 2023C03053.

% \bibliographystyle{aaai24_bib_sty}
% \bibliography{aaai24}

\clearpage
\begin{figure}[ht]
    \centering
    \includegraphics[width=0.8\linewidth]{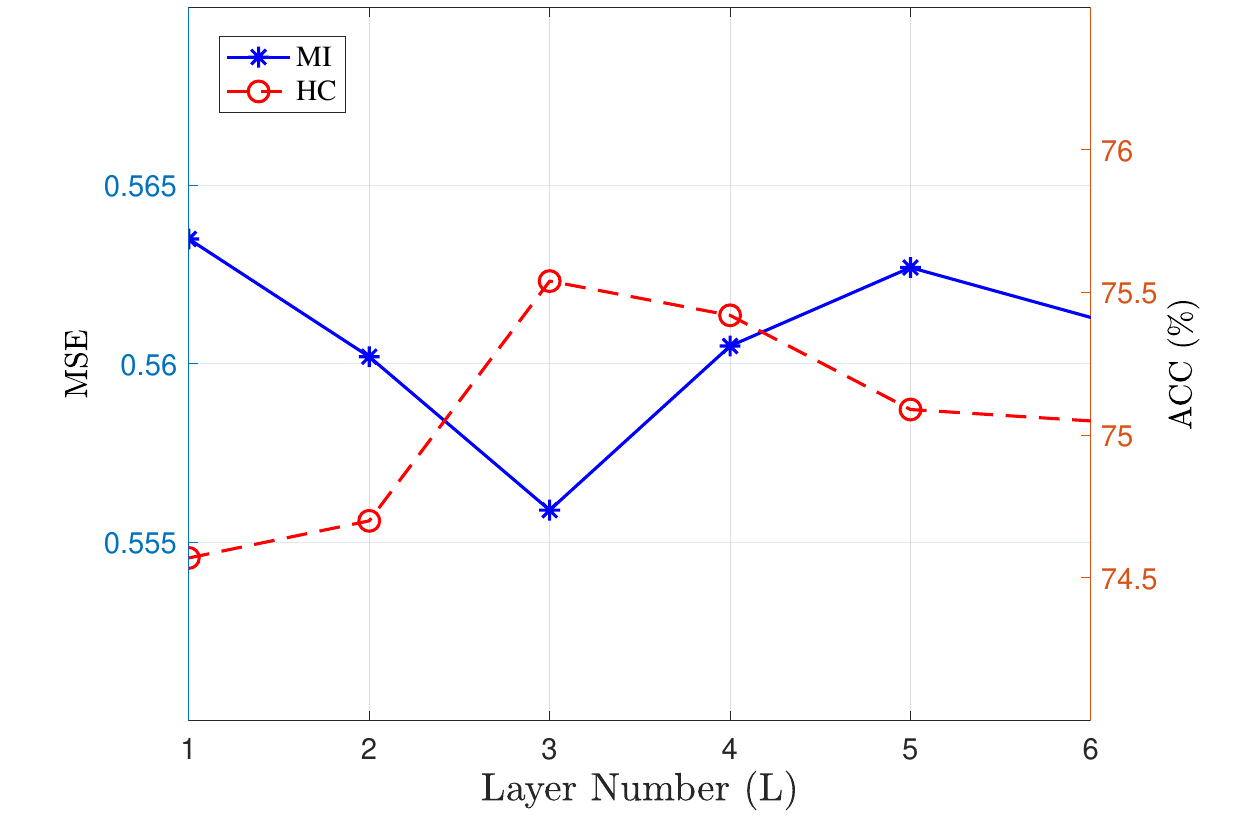}
    \caption{Impact of layer number L on MI and HC. 
    % Orange and blue rectangles on the right side are inputs and prompt tokens.
    }
    \label{fig: para L}
\end{figure}
\begin{figure}[ht]
    \centering
    \includegraphics[width=0.8\linewidth]{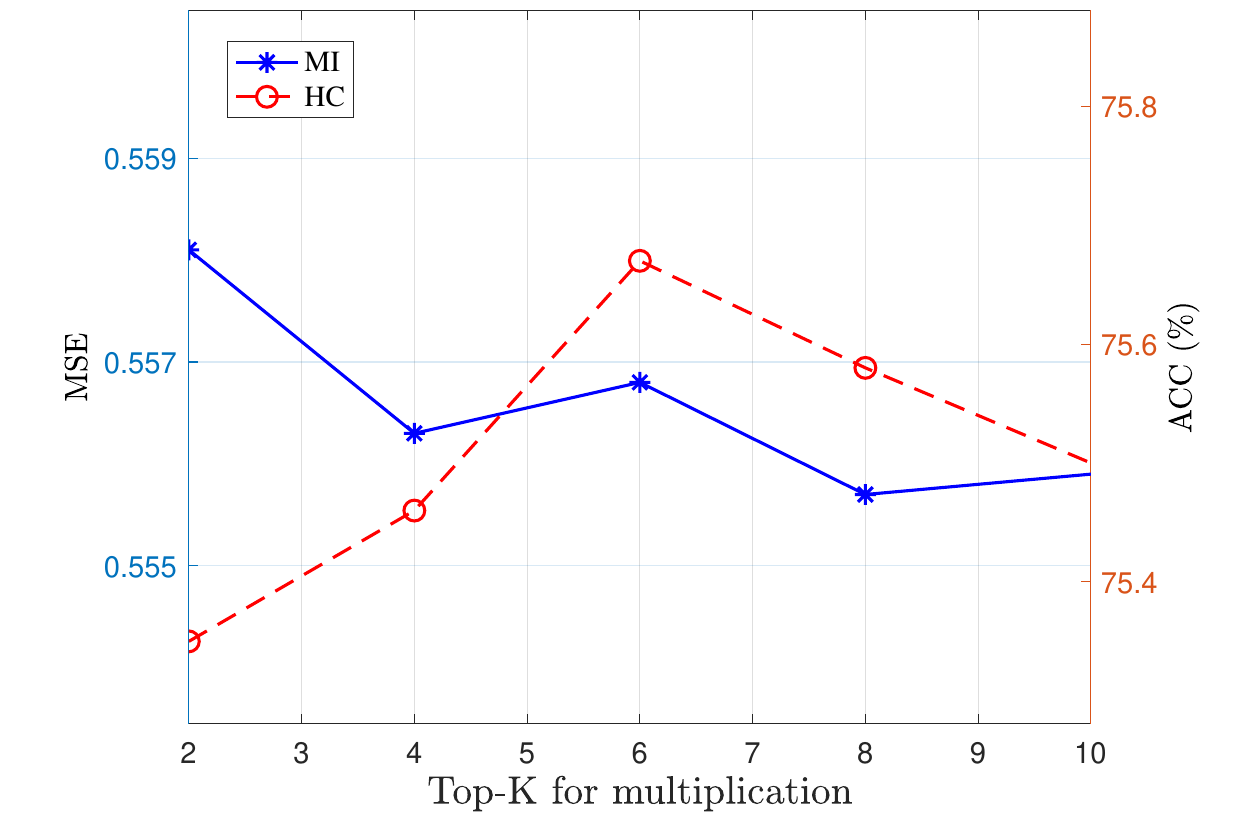}
    \caption{Parameter K on MI and HC. 
    % Orange and blue rectangles on the right side are inputs and prompt tokens.
    }
    \label{fig: para K}
\end{figure}
\section*{Appendix A: Extra Results on Parameter Sensitivity}
We also investigate the effect of the parameter $L$ on the MI and HC datasets, while keeping the default settings for other parameters. The MSE ($\downarrow$) and ACC ($\uparrow$) results are illustrated in Fig.~\ref{fig: para L}. Similar to the performance on the EP and CO datasets, as the number of layers increases, the overall performance of our \model initially improves and then starts to decline. The optimal performance is achieved on both datasets when $L=3$.

Considering the results obtained from all four datasets, it can be concluded that $L=3$ is a parameter that universally applies across different datasets.

With the optimal number of layers $L$, we further explore the impact of different values of $K$ on these two datasets. As depicted in Fig.~\ref{fig: para K}, the ACC on the HC dataset demonstrates a rapid increase before $K=6$ followed by a slight decline, while the MSE on the MI dataset continues to decrease and stabilizes after $K=8$. These findings suggest that a larger value of $K$ can enhance the performance of our \model, but increasing $K$ beyond a certain threshold may lead to a slight decline in performance.

\begin{table}[ht!]
    \centering
    \begin{tabular}{cc}
        % \toprule
        \hline          
        dimension & 1024 \\ 
        \# Layer & 2 \\
        choice-function & entmax15 \\
        bin-function & entmoid15 \\ 
        \hline
    \end{tabular}
    \caption{Default Settings for NODE.
    }
    \label{tab:setting node}
\end{table}

\begin{table}[ht!]
    \centering
    \begin{tabular}{cc}
        \hline          
        \# Layers & 3 \\ 
        dimension & 32 for AutoInt, 192 otherwise\\
        heads & 8       \\
        FF-dropout & 0.1       \\
        Attention-dropout & 0.2       \\
        \hline
    \end{tabular}
    \caption{Default Settings for Transformer-based methods, \ie FT-Transformer, DCAP and AutoInt.
    }
    \label{tab:setting transformer}
\end{table}

\begin{table}[ht!]
    \centering
    \begin{tabular}{cc}
        \hline          
        layer-num & 2 \\
        embedding-size & 16 \\
        dnn-hidden-units & (562, 562, 562) \\
        init-std & 0.0001 \\
        l2-reg & 0.00001 \\
        drop-rate & 0.5 \\
        \hline
    \end{tabular}
    \caption{Default Settings for DCN-V2.
    }
    \label{tab:setting dcnv2}
\end{table}

\begin{table}[ht!]
    \centering
    \begin{tabular}{cc}
        % \toprule
        \hline          
        booster & "gbtree" \\ 
        early-stopping-rounds & 50 \\
        n-estimators & 2000       \\
        \hline
    \end{tabular}
    \caption{Default Settings for XGBoost.
    }
    \label{tab:setting xgboost}
\end{table}

\section*{Appendix B: Default Settings for Each Method}
In this section, we provide the implementation details for all models. We utilize the official implementation for NODE~\footnote{https://github.com/Qwicen/node},  which uses the default hyper-parameter settings presented in Table~\ref{tab:setting node}. As shown Table~\ref{tab:setting transformer}, Transformer-based methods share most of the hyper-parameters, except for the feature dimension. AutoInt uses a feature dimension of 32, while FT-Transformer and DCAP use a feature dimension of 192. The hyper-parameters for DCN-V2 are listed in Table~\ref{tab:setting dcnv2}.It should be noted that the embedding feature within DCN-V2 is calculated as the sum of the number of numerical features and the product of the number of categorical features and the embedding size. Additionally, the default hyperparameter settings for XGBoost can be found in Table~\ref{tab:setting xgboost}.

\end{document}